\documentclass[10pt,twocolumn,letterpaper]{article}

\usepackage{xargs}                      
\usepackage[pdftex,dvipsnames]{xcolor}  
\usepackage[colorinlistoftodos,prependcaption,textsize=tiny]{todonotes}
\newcommandx{\unsure}[2][1=]{\todo[linecolor=red,backgroundcolor=red!25,bordercolor=red,#1]{#2}}
\newcommandx{\change}[2][1=]{\todo[linecolor=blue,backgroundcolor=blue!25,bordercolor=blue,#1]{#2}}
\newcommandx{\info}[2][1=]{\todo[linecolor=OliveGreen,backgroundcolor=OliveGreen!25,bordercolor=OliveGreen,#1]{#2}}
\newcommandx{\improvement}[2][1=]{\todo[linecolor=Plum,backgroundcolor=Plum!25,bordercolor=Plum,#1]{#2}}
\newcommandx{\thiswillnotshow}[2][1=]{\todo[disable,#1]{#2}}

\usepackage{fontawesome}
\usepackage{cvpr}
\usepackage{times}
\usepackage{epsfig}
\usepackage{graphicx}
\usepackage{amsmath}
\usepackage{amssymb}
\usepackage[nolist]{acronym}
\usepackage{color,soul}
\usepackage[justification=centering]{caption}
\usepackage[normalem]{ulem}
\usepackage{xspace}
\usepackage{pdfpages}
\usepackage{wasysym}
\usepackage{nopageno}

\DeclareSymbolFont{bbold}{U}{bbold}{m}{n}
\DeclareSymbolFontAlphabet{\mathbbold}{bbold}

\usepackage{tabularx}
\newcommand\setrow[1]{\gdef\rowmac{#1}#1\ignorespaces}
\newcommand\clearrow{\global\let\rowmac\relax}
\clearrow

\usepackage[breaklinks=true,bookmarks=false]{hyperref}

\cvprfinalcopy 

\def\cvprPaperID{289} 

\begin{acronym}[PHOENIX14\textbf{T} ]

\acrodefplural{slrt}[SLRTs]{Sign Language Recognition Transformers}
\acrodefplural{sltt}[SLTTs]{Sign Language Translation Transformers}

\acro{slrt}[SLRT]{Sign Language Recognition Transformer}
\acro{sltt}[SLTT]{Sign Language Translation Transformer}

\acrodefplural{rnn}[RNNs]{Recurrent Neural Networks}
\acrodefplural{cnn}[CNNs]{Convolutional Neural Networks}
\acrodefplural{hmm}[HMMs]{Hidden Markov Models}
\acrodefplural{gru}[GRUs]{Gated Recurrent Units}
\acrodefplural{crf}[CRFs]{Conditional Random Fields}
\acrodefplural{gan}[GANs]{Generative Adversarial Networks}
\acrodefplural{lcs}[LCSes]{Longest Common Subsequences}
\acrodefplural{gpu}[GPUs]{Graphic Processing Units}
\acro{bsl}[BSL]{British Sign Language}
\acro{bleu}[BLEU]{Bilingual Evaluation Understudy}
\acro{blstm}[BLSTM]{Bidirectional Long Short-Term Memory}
\acro{cnn}[CNN]{Convolutional Neural Network}
\acro{crf}[CRF]{Conditional Random Field}
\acro{cslr}[CSLR]{Continuous Sign Language Recognition}
\acro{ctc}[CTC]{Connectionist Temporal Classification}
\acro{dl}[DL]{Deep Learning}
\acro{dgs}[DGS]{German Sign Language - Deutsche Gebärdensprache}
\acro{dsgs}[DSGS]{Swiss German Sign Language - Deutschschweizer Geb\"ardensprache}
\acro{fc}[FC]{Fully Connected}
\acro{gan}[GAN]{Generative Adversarial Network}
\acro{gpu}[GPU]{Graphics Processing Unit}
\acro{gru}[GRU]{Gated Recurrent Unit}
\acro{hmm}[HMM]{Hidden Markov Model}
\acro{hog}[HOG]{Histograms of Oriented Gradients}
\acro{ip}[IP]{Inner Product}
\acro{isl}[ISL]{Irish Sign Language}
\acro{lcs}[LCS]{Longest Common Subsequence}
\acro{lstm}[LSTM]{Long Short-Term Memory}
\acro{nmt}[NMT]{Neural Machine Translation}
\acro{ocr}[OCR]{Optical Character Recognition}
\acro{ph12}[PHOENIX12]{RWTH-PHOENIX-Weather-2012}
\acro{ph14}[PHOENIX14]{RWTH-PHOENIX-Weather-2014}
\acro{ph14t}[PHOENIX14\textbf{T}]{RWTH-PHOENIX-Weather-2014\textbf{T}}
\acro{relu}[RELU]{Rectified Linear Units}
\acro{rnn}[RNN]{Recurrent Neural Network}
\acro{rouge}[ROUGE]{Recall-Oriented Understudy for Gisting Evaluation}
\acro{sgd}[SGD]{Stochastic Gradient Descent}
\acro{sla}[SLA]{Sign Language Assessment}
\acro{slr}[SLR]{Sign Language Recognition}
\acro{slt}[SLT]{Sign Language Translation}
\acro{sift}[SIFT]{Scale Invariant Feature Transform}
\acro{surf}[SURF]{Speeded Up Robust Features}
\acro{wer}[WER]{Word Error Rate}

\end{acronym}

\begin{document}

\title{Sign Language Transformers: \\Joint End-to-end Sign Language Recognition and Translation\vspace{-0.5cm}}
\author{\parbox{16cm}{\centering
    {\large 
        Necati Cihan Camg\"{o}z{\small \kern 0.1em \textsuperscript{\faGraduationCap}}, 
        Oscar Koller{\small \kern 0.1em \textsuperscript{\faWindows}},
        Simon Hadfield{\small \kern 0.1em \textsuperscript{\faGraduationCap}} and 
        Richard Bowden{\small \kern 0.1em \textsuperscript{\faGraduationCap}}
    } \\
    {\normalsize 
        {\scriptsize \textsuperscript{\faGraduationCap}}CVSSP, University of Surrey, Guildford, UK,
        {\scriptsize \textsuperscript{\faWindows}}Microsoft, Munich, Germany
    } \\
    { 
        \tt\small\{n.camgoz, s.hadfield, r.bowden\}@surrey.ac.uk},
        \tt\small{oscar.koller@microsoft.com}
    }
}
\vspace{-5cm}

\maketitle

\maketitle
\begin{abstract}
\vspace{-0.1cm}
Prior work on Sign Language Translation has shown that having a mid-level sign gloss representation (effectively \emph{recognizing} the individual signs) improves the \emph{translation} performance drastically. In fact, the current state-of-the-art in translation requires gloss level tokenization in order to work. We introduce a novel transformer based architecture that jointly learns Continuous Sign Language Recognition and Translation while being trainable in an end-to-end manner. This is achieved by using a \ac{ctc} loss to bind the recognition and translation problems into a single unified architecture. This joint approach does not require any ground-truth timing information, simultaneously solving two co-dependant sequence-to-sequence learning problems and leads to significant performance gains.

We evaluate the recognition and translation performances of our approaches on the challenging \ac{ph14t} dataset. We report state-of-the-art sign language recognition and translation results achieved by our Sign Language Transformers. Our translation networks outperform both sign video to spoken language and gloss to spoken language translation models, in some cases more than doubling the performance (9.58 vs. 21.80 BLEU-4 Score). We also share new baseline translation results using transformer networks for several other text-to-text sign language translation tasks.
\end{abstract}
\vspace{-0.2in}
\section{Introduction}
Sign Languages are the native languages of the Deaf and their main medium of communication. As visual languages, they utilize multiple complementary channels\footnote{Linguists refer to these channels as articulators.} to convey information \cite{sutton1999linguistics}. This includes manual features, such as hand shape, movement and pose as well as non-manuals features, such as facial expression, mouth and movement of the head, shoulders and torso \cite{boyes2001hands}. 

The goal of sign language translation is to either convert written language into a video of sign (production) \cite{stoll2018sign,stoll2020text2sign} or to extract an equivalent spoken language sentence from a video of someone performing continuous sign \cite{camgoz2018neural}. However, in the field of computer vision, much of this latter work has focused on recognising the sequence of sign glosses\footnote{Sign glosses are spoken language words that match the meaning of signs and, linguistically, manifest as minimal lexical items.} (\ac{cslr}) rather than the full translation to a spoken language equivalent (\ac{slt}). This distinction is important as the grammar of sign and spoken languages are very different. These differences include (to name a few): different word ordering, multiple channels used to convey concurrent information and the use of direction and space to convey the relationships between objects. Put simply, the mapping between speech and sign is complex and there is no simple word-to-sign mapping.

\begin{figure}[!t]
\begin{center}
\vspace{-0.1in}
\includegraphics[trim={0.8cm 0.7cm 0.8cm 0.5cm},clip,width=0.85\linewidth]{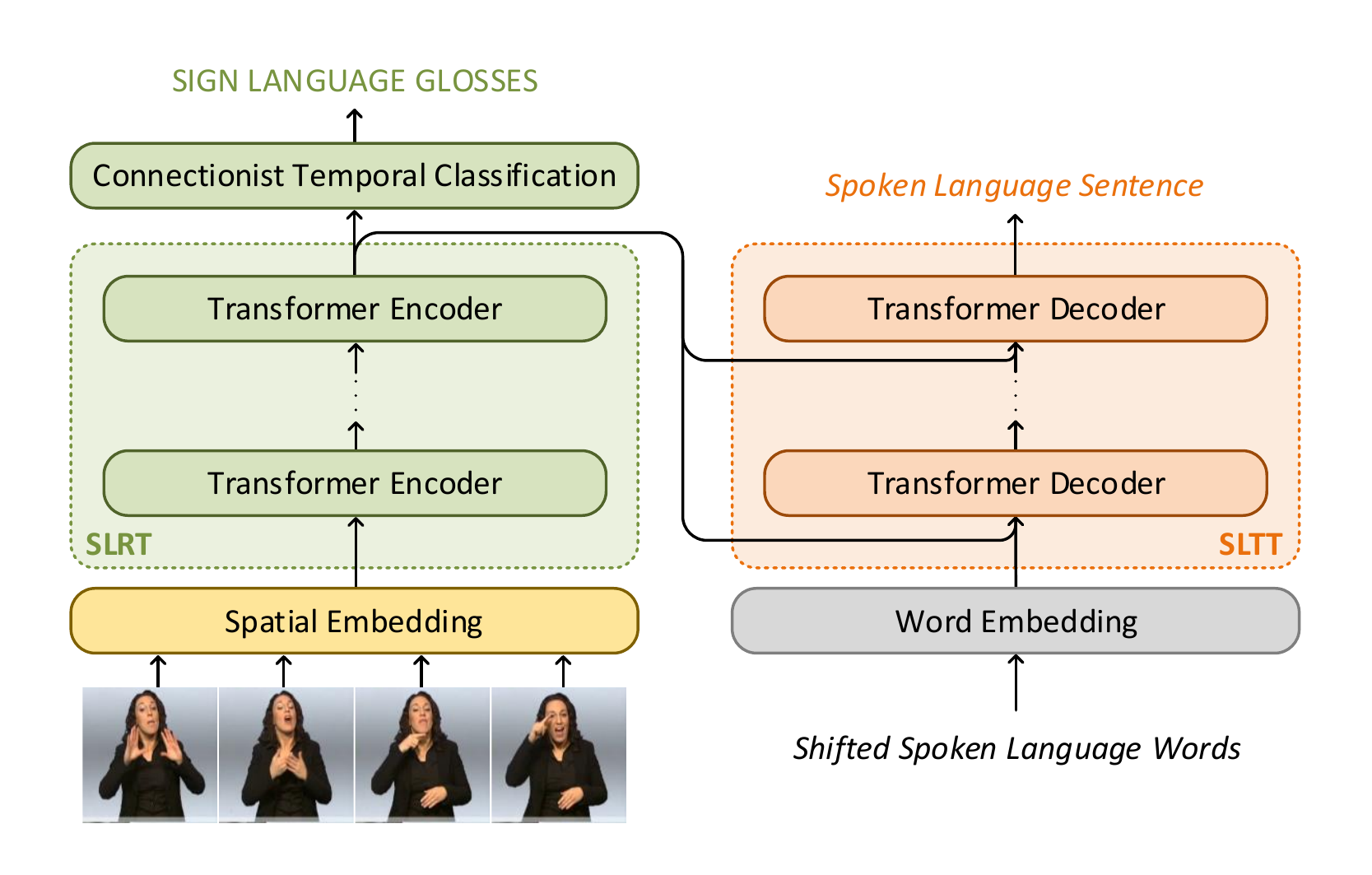}
\end{center}
\vspace{-0.2in}
\caption{An overview of our end-to-end Sign Language Recognition and Translation approach using transformers.}
\label{fig:overview}
\vspace{-0.2in}
\end{figure}

Generating spoken language sentences given sign language videos is therefore a spatio-temporal machine translation task \cite{camgoz2018neural}. Such a translation system requires us to accomplish several sub-tasks, which are currently unsolved:

\textbf{Sign Segmentation:} Firstly, the system needs to detect sign sentences, which are commonly formed using topic-comment structures \cite{sutton1999linguistics}, from continuous sign language videos. This is trivial to achieve for text based machine translation tasks \cite{neubig2017neural}, where the models can use punctuation marks to separate sentences. Speech-based recognition and translation systems, on the other hand, look for pauses, \eg silent regions, between phonemes to segment spoken language utterances \cite{van1991automatic,yoshida2002automatic}. There have been studies in the literature addressing automatic sign segmentation \cite{khan2014pause, santemiz2009automatic,shipman2017speed,borg2019sign,cherniavsky2008activity}. However to the best of the authors' knowledge, there is no study which utilizes sign segmentation for realizing continuous sign language translation.

\textbf{Sign Language Recognition and Understanding:} Following successful segmentation, the system needs to understand what information is being conveyed within a sign sentence. Current approaches tackle this by recognizing sign glosses and other linguistic components. Such methods can be grouped under the banner of \ac{cslr} \cite{koller2015continuous,camgoz2017subunets}. From a computer vision perspective, this is the most challenging task. Considering the input of the system is high dimensional spatio-temporal data, \ie sign videos, models are required that understand what a signer looks like and how they interact and move within their 3D signing space. Moreover, the model needs to comprehend what these aspects mean in combination. This complex modelling problem is exacerbated by the asynchronous multi-articulatory nature of sign languages \cite{sandler1993sign,stokoe1980sign}. Although there have been promising results towards \ac{cslr}, the state-of-the-art \cite{koller2019weakly} can only recognize sign glosses and operate within a limited domain of discourse, namely weather forecasts \cite{forster2014extensions}.  

\textbf{Sign Language Translation:} Once the information embedded in the sign sentences is understood by the system, the final step is to generate spoken language sentences. As with any other natural language, sign languages have their own unique linguistic and grammatical structures, which often do not have a one-to-one mapping to their spoken language counterparts. As such, this problem truly represents a machine translation task. Initial studies conducted by computational linguists have used text-to-text statistical machine translation models to learn the mapping between sign glosses and their spoken language translations \cite{morrissey2008data}. However, glosses are simplified representations of sign languages and linguists are yet to come to a consensus on how sign languages should be annotated. 

There have been few contributions towards video based continuous \ac{slt}, mainly due to the lack of suitable datasets to train such models. More recently, Camgoz \etal \cite{camgoz2018neural} released the first publicly available sign language video to spoken language translation dataset, namely \ac{ph14t}. In their work, the authors proposed approaching \ac{slt} as a \ac{nmt} problem. Using attention-based \ac{nmt} models \cite{luong2015effective,bahdanau2015neural}, they define several \ac{slt} tasks and realized the first end-to-end sign language video to spoken language sentence translation model, namely \emph{Sign2Text}. 

One of the main findings of \cite{camgoz2018neural} was that using gloss based mid-level representations improved the \ac{slt} performance drastically when compared to an end-to-end \emph{Sign2Text} approach. The resulting \emph{Sign2Gloss2Text} model first recognized  glosses from continuous sign videos using a state-of-the-art \ac{cslr} method \cite{koller2017resign}, which worked as a tokenization layer. The recognized sign glosses were then passed to a text-to-text attention-based \ac{nmt} network \cite{luong2015effective} to generate spoken language sentences.

We hypothesize that there are two main reasons why \emph{Sign2Gloss2Text} performs better than \emph{Sign2Text} (18.13 vs 9.58 BLEU-4 scores). Firstly, the number of sign glosses is much lower than the number of frames in the videos they represent. By using gloss representations instead of the spatial embeddings extracted from the video frames, \emph{Sign2Gloss2Text} avoids the long-term dependency issues, which \emph{Sign2Text} suffers from.

We think the second and more critical reason is the lack of direct guidance for understanding sign sentences in \emph{Sign2Text} training. Given the aforementioned complexity of the task, it might be too difficult for current Neural Sign Language Translation architectures to comprehend sign without any explicit intermediate supervision. In this paper we propose a novel Sign Language Transformer approach, which addresses this issue while avoiding the need for a two-step pipeline, where translation is solely dependent on recognition accuracy. This is achieved by jointly learning sign language recognition and translation from spatial-representations of sign language videos in an end-to-end manner. Exploiting the encoder-decoder based architecture of transformer networks \cite{vaswani2017attention}, we propose a multi-task formalization of the joint continuous sign language recognition and translation problem.

To help our translation networks with sign language understanding and to achieve \ac{cslr}, we introduce a \ac{slrt}, an encoder transformer model trained using a \ac{ctc} loss \cite{afouras2018deep}, to predict sign gloss sequences. \ac{slrt} takes spatial embeddings extracted from sign videos and learns spatio-temporal representations. These representations are then fed to the \ac{sltt}, an autoregressive transformer decoder model, which is trained to predict one word at a time to generate the corresponding spoken language sentence. An overview of the approach can be seen in Figure~\ref{fig:overview}.

The contributions of this paper can be summarized as:
\begin{itemize}
    \vspace{-0.2cm}
    \item A novel multi-task formalization of \ac{cslr} and \ac{slt} which exploits the supervision power of glosses, without limiting the translation to spoken language.
    \vspace{-0.25cm}
    \item The first successful application of transformers for \ac{cslr} and \ac{slt} which achieves state-of-the-art results in both recognition and translation accuracy, vastly outperforming all comparable previous approaches.
    \vspace{-0.25cm}
    \item A broad range of new baseline results to guide future research in this field.
    \vspace{-0.2cm}
\end{itemize}

The rest of this paper is organized as follows: In Section~\ref{sec:related:work}, we survey the previous studies on \ac{slt} and the state-of-the-art in the field of \ac{nmt}. In Section~\ref{sec:methodology}, we introduce Sign Language Transformers, a novel joint sign language recognition and translation approach which can be trained in an end-to-end manner. We share our experimental setup in Section~\ref{sec:dataset:tasks}. We then report quantitative results of the Sign Language Transformers in Section~\ref{sec:quant} and present new baseline results for the previously defined text-to-text translation tasks \cite{camgoz2018neural}. In Section~\ref{sec:qual}, we share translation examples generated by our network to give the reader further qualitative insight of how our approach performs. We conclude the paper in Section \ref{sec:conc} by discussing our findings and possible future work. 

\section{Related Work}
\label{sec:related:work}
\vspace{-0.2cm}
Sign languages have been studied by the computer vision community for the last three decades \cite{tamura1988recognition,starner1998real}. The end goal of computational sign language research is to build translation and production systems \cite{cormier2019extol}, that are capable of translating sign language videos to spoken language sentences and vice versa, to ease the daily lives of the Deaf \cite{cooper2011sign,bragg2019sign}. However, most of the research to date has mainly focused on Isolated Sign Language Recognition \cite{joze2019ms, yin2016iterative,wang2016grassman,camgoz2016sign,suzgun2015hospisign,tornay2019hmm}, working on application specific datasets \cite{camgoz2016bosphorussign,wang2016sparse,ebling2018smile}, thus limiting the applicability of such technologies. More recent work has tackled continuous data \cite{koller2016deepsign,huang2018video,cui2017recurrent,cui2019deep}, but the move from recognition to translation is still in its infancy \cite{camgoz2018neural}.\looseness=-1

There have been earlier attempts to realize \ac{slt} by computational linguists. However, existing 
work has solely focused on the text-to-text translation problem and has been very limited in size, averaging around 3000 total words \cite{morrissey2010building, stein2012analysis, schmid2013using}. Using statistical machine translation methods, Stein \etal \cite{stein2012analysis} proposed a weather broadcast translation system from spoken German into \ac{dgs} and vice versa, using the \ac{ph12} \cite{forster2012rwth} dataset. Another method translated air travel information from spoken English to \ac{isl}, spoken German to \ac{isl}, spoken English to \ac{dgs}, and spoken German to \ac{dgs} \cite{morrissey2008data}. Ebling \cite{ebling2016Automatic} developed an approach to translate written German train announcements into \ac{dsgs}. While non-manual information has not been included in most previous systems, Ebling \& Huenerfauth \cite{ebling2015bridging} proposed a sequence classification based model to schedule the automatic generation of non-manual features after the core machine translation step. 

Conceptual video based \ac{slt} systems were introduced in the early 2000s \cite{bungeroth2004statistical}. There have been studies, such as \cite{chai2013sign}, which propose recognizing signs in isolation and then constructing sentences using a language model. However, end-to-end \ac{slt} from video has not been realized until recently. 

The most important obstacle to vision based \ac{slt} research has been the availability of suitable datasets. Curating and annotating continuous sign language videos with spoken language translations is a laborious task. There are datasets available from linguistic sources \cite{schembri2013building, hanke2010dgs} and sign language interpretations from broadcasts \cite{cooper2009learning}. However, the available annotations are either weak (subtitles) or too few to build models which would work on a large domain of discourse. In addition, such datasets lack the human pose information which legacy \ac{slr} methods heavily relied on.

The relationship between sign sentences and their spoken language translations are non-monotonic, as they have different ordering. Also, sign glosses and linguistic constructs do not necessarily have a one-to-one mapping with their spoken language counterparts. This made the use of available \ac{cslr} methods \cite{koller2016deepsign,koller2017resign} (that were designed to learn from weakly annotated data) infeasible, as they are build on the assumption that sign language videos and corresponding annotations share
the same temporal order.

To address these issues, Camgoz \etal \cite{camgoz2018neural} released the first publicly available \ac{slt} dataset, \ac{ph14t}, which is an extension of the popular \ac{ph14} \ac{cslr} dataset. The authors approached the task as a spatio-temporal neural machine translation problem, which they term \emph{`Neural Sign Language Translation'}. They proposed a system using \acp{cnn} in combination with attention-based \ac{nmt} methods \cite{luong2015effective, bahdanau2015neural} to realize the first end-to-end \ac{slt} models. Following this, Ko \etal proposed a similar approach but used body key-point coordinates as input for their translation networks, and evaluated their method on a Korean Sign Language dataset \cite{ko2019neural}. 

Concurrently, there have been several advancements in the field of \ac{nmt}, one of the most important being the introduction of transformer networks \cite{vaswani2017attention}. Transformers drastically improved the translation performance over legacy attention based encoder-decoder approaches. Also due to the fully-connected nature of the architecture, transformers are fast and easy to parallelize, which has enabled them to become the new go to architecture for many machine translation tasks. In addition to \ac{nmt}, transformers have achieved success in various other challenging tasks, such as language modelling \cite{dai2019transformer,zhang2019ernie}, learning sentence representations \cite{devlin2018bert}, multi-modal language understanding \cite{tsai2019MULT}, activity \cite{wang2018non} and speech recognition \cite{irie2019language}. Inspired by their recent wide-spread success, in this work we propose a novel architecture where multiple co-dependent transformer networks are simultaneously trained to jointly solve related tasks. We then apply this architecture to the problem of simultaneous recognition and translation where joint training provides significant benefits. 

\begin{figure*}[!ht]
\vspace{-0.25in}
\begin{center}
\includegraphics[trim={1.6cm 0.4cm 1.1cm 0.3cm},clip,width=0.78\linewidth]{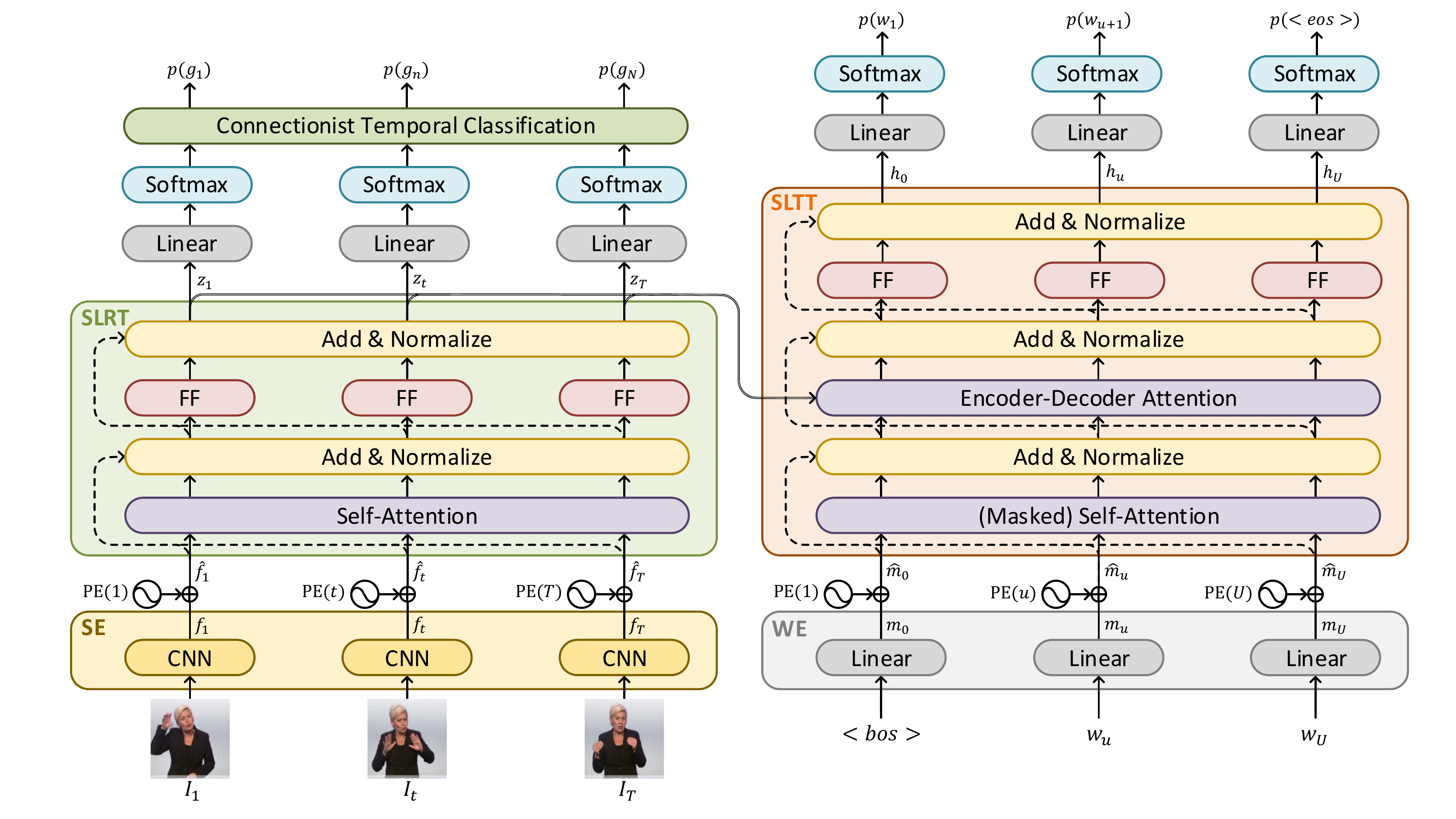}
\end{center}
\vspace{-0.15in}
\caption{A detailed overview of a single layered Sign Language Transformer. \\ (SE: Spatial Embedding, WE: Word Embedding , PE: Positional Encoding, FF: Feed Forward)}
\label{fig:overview:detailed}
\vspace{-0.10in}
\end{figure*}
\section{Sign Language Transformers}
\label{sec:methodology}
In this section we introduce Sign Language Transformers which jointly learn to recognize and translate sign video sequences into sign glosses and spoken language sentences in an end-to-end manner. Our objective is to learn the conditional probabilities $p(\mathcal{G}|\mathcal{V})$ and $p(\mathcal{S}|\mathcal{V})$ of generating a sign gloss sequence $\mathcal{G} = (g_1, ..., g_N)$ with $N$ glosses and a spoken language sentence $\mathcal{S} = (w_1, ..., w_U)$ with $U$ words given a sign video $\mathcal{V} = (I_1, ..., I_T)$ with $T$ frames.

Modelling these conditional probabilities is a sequence-to-sequence task, and poses several challenges. In both cases, the number of tokens in the source domain is much larger than the corresponding target sequence lengths (\ie $T\gg{N}$ and $T\gg{U}$). Furthermore, the mapping between sign language videos, $\mathcal{V}$, and spoken language sentences, $\mathcal{S}$, is non-monotonic, as both languages have different vocabularies, grammatical rules and orderings. 

Previous sequence-to-sequence based literature on \ac{slt} can be categorized into two groups: The first group break down the problem in two stages. They consider \ac{cslr} as an initial process and then try to solve the problem as a text-to-text translation task \cite{chai2013sign, camgoz2018neural}. Camgoz \etal utilized a state-of-the-art \ac{cslr} method \cite{koller2017resign} to obtain sign glosses, and then used an attention-based text-to-text \ac{nmt} model \cite{luong2015effective} to learn the sign gloss to spoken language sentence translation, $p(\mathcal{S}|\mathcal{G})$ \cite{camgoz2018neural}. However, in doing so, this approach introduces an information bottleneck in the mid-level gloss representation. This limits the network's ability to understand sign language as the translation model can only be as good as the sign gloss annotations it was trained from. There is also an inherent loss of information as a sign gloss is an incomplete annotation intended only for linguistic study and it therefore neglects many crucial details and information present in the original sign language video.

The second group of methods focus on translation from the sign video representations to spoken language with no intermediate representation \cite{camgoz2018neural, ko2019neural}. These approaches attempt to learn $p(\mathcal{S}|\mathcal{V})$ directly. Given enough data and a sufficiently sophisticated network architecture, such models could theoretically realize end-to-end \ac{slt} with no need for a human-interpretable information that act as a bottleneck. However, due to the lack of direct supervision guiding sign language understanding, such methods have significantly lower performance than their counterparts on currently available datasets \cite{camgoz2018neural}. 

To address this, we propose to jointly learn $p(\mathcal{G}|\mathcal{V})$ and $p(\mathcal{S}|\mathcal{V})$, in an end-to-end manner. We build upon transformer networks \cite{vaswani2017attention} to create a unified model, which we call Sign Language Transformers (See Figure~\ref{fig:overview:detailed}). We train our networks to generate spoken language sentences from sign language video representations. During training, we inject intermediate gloss supervision in the form of a \ac{ctc} loss into the \acf{slrt} encoder. This helps our networks learn more meaningful spatio-temporal representations of the sign without limiting the information passed to the decoder. We employ an autoregressive \acf{sltt} decoder which predicts one word at a time to generate the spoken language sentence translation.

\subsection{Spatial and Word Embeddings}
Following the classic \ac{nmt} pipeline, we start by embedding our source and target tokens, namely sign language video frames and spoken language words. As word embedding we use a linear layer, which is initialized from scratch during training, to project a one-hot-vector representation of the words into a denser space. To embed video frames, we use the SpatialEmbedding approach \cite{camgoz2018neural}, and propagate each image through \acp{cnn}. We formulate these operations as:\looseness=-1
\begin{align}
\label{eqn:eqlabel}
\begin{split}
    m_u &= \mathrm{WordEmbedding}(w_u) \\
    f_t &= \mathrm{SpatialEmbedding}(I_t)
\end{split}
\end{align}
where $m_u$ is the embedded representation of the spoken language word $w_u$ and $f_t$ corresponds to the non-linear frame level spatial representation obtained from a \ac{cnn}.

Unlike other sequence-to-sequence models \cite{sutskever2014sequence,gehring2017convolutional}, transformer networks do not employ recurrence or convolutions, thus lacking the positional information within sequences. To address this issue we follow the positional encoding method proposed in \cite{vaswani2017attention} and add temporal ordering information to our embedded representations as:
\begin{alignat*}{2}
    \hat{f}_{t} &= f_t &&+ \mathrm{PositionalEncoding}(t) \\
    \hat{m}_{u} &= m_u &&+ \mathrm{PositionalEncoding}(u)
\end{alignat*}
where $\mathrm{PositionalEncoding}$ is a predefined function which produces a unique vector in the form of a phase shifted sine wave for each time step. 

\subsection{Sign Language Recognition Transformers}
The aim of \ac{slrt} is to recognize glosses from continuous sign language videos while learning meaningful spatio-temporal representations for the end goal of sign language translation. Using the positionally encoded spatial embeddings, $\hat{f}_{1:T}$, we train a transformer encoder model \cite{vaswani2017attention}.

The inputs to \ac{slrt} are first modelled by a Self-Attention layer which learns the contextual relationship between the frame representations of a video. Outputs of the self-attention are then passed through a non-linear point-wise feed forward layer. All the operations are followed by residual connections and normalization to help training. We formulate this encoding process as:
\vspace{-0.2cm}
\begin{equation}
\vspace{-0.2cm}
    z_t = \mathrm{SLRT}(\hat{f}_{t}|\hat{f}_{1:T})
\end{equation}
where $z_t$ denotes the spatio-temporal representation of the frame $I_t$, which is generated by \ac{slrt} at time step $t$, given the spatial representations of all of the video frames, $\hat{f}_{1:T}$. 

We inject intermediate supervision to help our networks understand sign and to guide them to learn a meaningful sign representation which helps with the main task of translation. We train the \ac{slrt} to model $p(\mathcal{G}|\mathcal{V})$ and predict sign glosses. Due to the spatio-temporal nature of the signs, glosses have a one-to-many mapping to video frames but share the same ordering.

One way to train the \ac{slrt} would be using cross-entropy loss \cite{goodfellow2016deep} with frame level annotations. However, sign gloss annotations with such precision are rare. An alternative form of weaker supervision is to use a sequence-to-sequence learning loss functions, such as \ac{ctc} \cite{graves2006connectionist}. 

Given spatio-temporal representations, $z_{1:T}$, we obtain frame level gloss probabilities, $p(g_t | \mathcal{V})$, using a linear projection layer followed by a softmax activation. We then use \ac{ctc} to compute  $p(\mathcal{G}|\mathcal{V})$ by marginalizing over all possible $\mathcal{V}$ to $\mathcal{G}$ alignments as: 
\vspace{-0.2cm}
\begin{equation}
\vspace{-0.2cm}
    p(\mathcal{G}|\mathcal{V}) = \sum_{\pi \in \mathcal{B}} p (\pi | \mathcal{V})  
\end{equation}
where $\pi$ is a path and $\mathcal{B}$ are the set of all viable paths that correspond to $\mathcal{G}$. We then use the $p(\mathcal{G}|\mathcal{V})$ to calculate the \ac{cslr} loss as:
\begin{equation}
    \mathcal{L}_\mathrm{R} =  1 - p(\mathcal{G^*}|\mathcal{V})
\end{equation}
where $\mathcal{G^*}$ is the ground truth gloss sequence.

\subsection{Sign Language Translation Transformers}
The end goal of our approach is to generate spoken language sentences from sign video representations. We propose training an autoregressive transformer decoder model, named \ac{sltt}, which exploits the spatio-temporal representations learned by the \ac{slrt}. 
We start by prefixing the target spoken language sentence $\mathcal{S}$ with the special beginning of sentence token, $\mathrm{<bos>}$. We then extract the positionally encoded word embeddings. These embeddings are passed to a masked self-attention layer. Although the main idea behind self-attention is the same as in \ac{slrt}, the \ac{sltt} utilizes a mask over the self-attention layer inputs. This ensures that each token may only use its predecessors while extracting contextual information. This masking operation is necessary, as at inference time the \ac{sltt} won't have access to the output tokens which would follow the token currently being decoded.

Representations extracted from both \ac{slrt} and \ac{sltt} self-attention layers are combined and given to an encoder-decoder attention module which learns the mapping between source and target sequences. Outputs of the encoder-decoder attention are then passed through a non-linear point-wise feed forward layer. Similar to \ac{slrt}, all the operations are followed by residual connections and normalization. We formulate this decoding process as:\looseness=-1
\vspace{-0.10cm}
\begin{equation}
    \vspace{-0.10cm}
    h_{u+1} = \mathrm{SLTT}(\hat{m}_{u}|\hat{m}_{1:u-1},z_{1:T}).
\end{equation}
\ac{sltt} learns to generate one word at a time until it produces the special end of sentence token, $\mathrm{<eos>}$. It is trained by decomposing the sequence level conditional probability $p(\mathcal{S|V})$ into ordered conditional probabilities
\vspace{-0.10cm}
\begin{equation}
    \vspace{-0.10cm}
	p(\mathcal{S|V}) = \prod_{u=1}^{U} p(w_u|h_u) 
\end{equation}
which are used to calculate the cross-entropy loss for each word as:
\vspace{-0.10cm}
\begin{equation}
    \vspace{-0.10cm}
    \mathcal{L}_T = 1 - \prod_{u=1}^{U} \sum_{d=1}^D p(\hat{w}^d_u) p(w^d_u|h_u) 
\end{equation}
where $p(\hat{w}^d_u)$ represents the ground truth probability of word $w^d$ at decoding step $u$ and $D$ is the target language vocabulary size.

We train our networks by minimizing the joint loss term $\mathcal{L}$, which is a weighted sum of the recognition loss $\mathcal{L}_\mathrm{R}$ and the translation loss $\mathcal{L}_\mathrm{T}$ as:
\begin{equation}
	\mathcal{L} = \lambda_\mathrm{R} \mathcal{L}_\mathrm{R} + \lambda_\mathrm{T} \mathcal{L}_\mathrm{T}
\end{equation}
where $\lambda_{R}$ and $\lambda_{T}$ are hyper parameters which decides the importance of each loss function during training and are evaluated in Section~\ref{sec:quant}.

\section{Dataset and Translation Protocols}
\label{sec:dataset:tasks}

We evaluate our approach on the recently released \ac{ph14t} dataset \cite{camgoz2018neural}, which is a large vocabulary, continuous \ac{slt} corpus. \ac{ph14t} is a translation focused extension of the \ac{ph14} corpus, which has become the primary benchmark for \ac{cslr} in recent years.

\ac{ph14t} contains parallel sign language videos, gloss annotations and their translations, which makes it the only available dataset suitable for training and evaluating joint SLR and SLT techniques. The corpus includes unconstrained continuous sign language from 9 different signers with a vocabulary of 1066 different signs. Translations for these videos are provided in German spoken language with a vocabulary of 2887 different words.

The evaluation protocols on the \ac{ph14t} dataset, as laid down by \cite{camgoz2018neural}, are as follows:

\textbf{\emph{Sign2Text}} is the end goal of \ac{slt}, where the objective is to translate directly from continuous sign videos to spoken language sentences without going via any intermediary representation, such as glosses.

\textbf{\emph{Gloss2Text}} is a text-to-text translation problem, where the objective is to translate ground truth sign gloss sequences to German spoken language sentences. The results of these experiments act as a virtual upper bound for the available \ac{nmt} translation technology. This assumption is based on the fact that perfect sign language recognition/understanding is simulated by using the ground truth gloss annotation. However, as mentioned earlier, one needs to bear in mind that gloss representations are imprecise. As glosses are textual representations of multi-channel temporal signals, they represent an information bottleneck for any translation system. This means that under ideal conditions, a \emph{Sign2Text} system could and should outperform \emph{Gloss2Text}. However, more sophisticated network architectures and data are needed to achieve this and hence such a goal remains a longer term objective beyond the scope of this manuscript. 

\textbf{\emph{Sign2Gloss2Text}} is the current state-of-the-art in \ac{slt}. This approach utilizes \ac{cslr} models to extract gloss sequences from sign language videos which are then used to solve the translation task as a text-to-text problem by training a \emph{Gloss2Text} network using the \ac{cslr} predictions.

\textbf{\emph{Sign2Gloss$\rightarrow$Gloss2Text}} is similar to \emph{Sign2Gloss2Text} and also uses \ac{cslr} models to extract gloss sequences. However, instead of training text-to-text translation networks from scratch, \emph{Sign2Gloss$\rightarrow$Gloss2Text} models use the best performing \emph{Gloss2Text} network, which has been trained with ground truth gloss annotations, to generate spoken language sentences from intermediate sign gloss sequences from the output of the \ac{cslr} models.

In addition to evaluating our networks in the context of the above protocols, we additionally introduce two new protocols which follow the same naming convention. \textbf{\emph{Sign2Gloss}} is a protocol which essentially performs \ac{cslr}, while \textbf{\emph{Sign2(Gloss+Text)}} requires joint learning of continuous sign language recognition and translation.

\section{Quantitative Results}
\label{sec:quant}
In this section we share our sign language recognition and translation experimental setups and report quantitative results. We first go over the implementation details and introduce the evaluation metrics we will be using to measure the performance of our models. We start our experiments by applying transformer networks to the text-to-text based \ac{slt} tasks, namely  \emph{Gloss2Text}, \emph{Sign2Gloss2Text}, \emph{Sign2Gloss$\rightarrow$Gloss2Text} and report improved performance over using \ac{rnn} based models. We share our \emph{Sign2Gloss} experiments, in which we explore the effects of different types of spatial embeddings and network structures on the performance of \ac{cslr}. We then train \emph{Sign2Text} and \emph{Sign2(Gloss+Text)} models using the best performing \emph{Sign2Gloss} configuration and investigate the effect of different recognition loss weights on the joint recognition and translation performance. Finally, we compare our best performing models against other approaches and report state-of-the-art results.

\begin{table*}[!h]
\centering
\vspace{-0.1in}
\resizebox{\linewidth}{!}{
\begin{tabular}{>{\rowmac}r|>{\rowmac}c|>{\rowmac}c>{\rowmac}c>{\rowmac}c>{\rowmac}c|>{\rowmac}c|>{\rowmac}c>{\rowmac}c>{\rowmac}c>{\rowmac}c}
   & \multicolumn{5}{c|}{DEV} &\multicolumn{5}{c}{TEST} \\
\hline Text-to-Text Tasks (RNNs vs Transformers) & WER & BLEU-1 & BLEU-2 & BLEU-3 & BLEU-4 & WER & BLEU-1 & BLEU-2 & BLEU-3 & BLEU-4\\
\hline	
Gloss2Text \cite{camgoz2018neural}  & - & 44.40 & 31.83 & 24.61 & 20.16 & - & 44.13 & 31.47 & 23.89 & 19.26 \\
\setrow{\bfseries} Our Gloss2Text & - & 50.69 & 38.16 & 30.53 & 25.35 & - & 48.90 & 36.88 & 29.45 & 24.54 \clearrow \\
\hline			
Sign2Gloss2Text \cite{camgoz2018neural} & - & 42.88 & 30.30 & 23.02 & 18.40 & - & 43.29 & 30.39 & 22.82 & 18.13 \\  
\setrow{\bfseries} Our Sign2Gloss2Text & - & 47.73 & 34.82 & 27.11 & \textbf{22.11} & - & 48.47 & 35.35 & 27.57 & 22.45 \clearrow \\
\hline
Sign2Gloss$\rightarrow$Gloss2Text \cite{camgoz2018neural} & - & 41.08 & 29.10 & 22.16 & 17.86 & - & 41.54 & 29.52 & 22.24 & 17.79 \\
\setrow{\bfseries} Our Sign2Gloss$\rightarrow$Gloss2Text & - & 47.84 & 34.65 & 26.88 & 21.84 & - & 47.74 & 34.37 & 26.55 & 21.59 \clearrow \\
\hline \hline			
Video-to-Text Tasks & WER & BLEU-1 & BLEU-2 & BLEU-3 & BLEU-4 & WER & BLEU-1 & BLEU-2 & BLEU-3 & BLEU-4\\
\hline 			
CNN+LSTM+HMM \cite{koller2019weakly} & 24.50 & - & - & - & - & 26.50 & - & - & - & -  \\
\setrow{\bfseries} Our Sign2Gloss & 24.88 & - & - & - & - & 24.59 & - & - & - & -   \clearrow \\
\hline
Sign2Text \cite{camgoz2018neural} & - & 31.87 & 19.11 & 13.16  & 9.94 & - & 32.24 & 19.03 & 12.83 & 9.58  \\
\setrow{\bfseries} Our Sign2Text & - & 45.54 &	32.60 &	25.30 &	20.69 & - & 45.34 &	32.31 &	24.83 &	20.17  \clearrow  \\
\hline \hline
\setrow{\bfseries} Our Best Recog. Sign2(Gloss+Text) & 24.61 & 46.56 & 34.03 & 26.83 & 22.12 & 24.49 & 47.20 & 34.46 & 26.75 & 21.80  \clearrow  \\
\setrow{\bfseries} Our Best Trans. Sign2(Gloss+Text) & 24.98 & 47.26 & 34.40 & 27.05 & 22.38 &  26.16 &	46.61 &	33.73 &	26.19 &	21.32 \clearrow \\
\noalign{\smallskip} 
\end{tabular}}
\vspace{-0.05in}
\caption{(Top) New baseline results for text-to-text tasks on Phoenix2014T \cite{camgoz2018neural} using transformer networks and \\ (Bottom) Our best performing Sign Language Transformers compared against the state-of-the-art.}
\label{tbl:ph14t:new:baselines} 
\vspace{-0.1in}
\end{table*}
\subsection{Implementation and Evaluation Details}
\textbf{Framework:} We used a modified version of JoeyNMT \cite{kreutzer2019joey} to implement our Sign Language Transformers\footnote{\url{https://github.com/neccam/slt}}. All components of our network were built using the PyTorch framework \cite{paszke2017automatic}, except the \ac{ctc} beam search decoding, for which we utilized the TensorFlow implementation \cite{abadi2016tensorflow}.

\textbf{Network Details:} Our transformers are built using $512$ hidden units and $8$ heads in each layer. We use Xavier initialization \cite{glorot2010understanding} and train all of our networks from scratch. We also utilize dropout with $0.1$ drop rate on transformer layers and word embeddings to mitigate over-fitting.

\textbf{Performance Metrics:}  We use Word Error Rate (WER) for assessing our recognition models, as it is the prevalent metric for evaluating \ac{cslr} performance \cite{koller2015continuous}. To measure the translation performance of our networks, we utilized BLEU \cite{papineni2002bleu} score (n-grams ranging from 1 to 4), which is the most common metric for machine translation. 

\textbf{Training:} We used the Adam \cite{kingma2014adam} optimizer to train our networks using a batch size of $32$ with a learning rate of $10^{-3}$ ($\beta_1$=$0.9$, $\beta_2$=$0.998$) and a weight decay of $10^{-3}$. We utilize plateau learning rate scheduling which tracks the development set performance. We evaluate our network every $100$ iterations. If the development score does not decrease for $8$ evaluation steps, we decrease the learning rate by a factor of $0.7$. This continues until the learning rate drops below $10^{-6}$.

\textbf{Decoding:} During the training and validation steps we employ a greedy search to decode both gloss sequences and spoken language sentences. At inference time, we utilize beam search decoding with widths ranging from $0$ to $10$. We also implement a length penalty \cite{wu2016google} with $\alpha$ values ranging from $0$ to $2$. We find the best performing combination of beam width and $\alpha$ on the development set and use these values for the test set evaluation.

\subsection{Text-to-Text Sign Language Translation}
\label{sec:text:baselines}
In our first set of experiments, we adapt the transformer backbone of our technique, for text-to-text sign language translation. We then evaluate the performance gain achieved over the \ac{rnn}-based attention architectures.

As can be seen in Table~\ref{tbl:ph14t:new:baselines}, utilizing transformers for text-to-text sign language translation improved the performance across all tasks, reaching an impressive $25.35$/$24.54$ ~{BLEU-4} score on the development and test sets. We believe this performance gain is due to the more sophisticated attention architectures, namely self-attention modules, which learn the contextual information within both source and target sequences. 

\subsection{\textbf{\emph{Sign2Gloss}}}

To tackle the \emph{Sign2Gloss} task, we utilize our \ac{slrt} networks. Any \ac{cnn} architecture can be used as spatial embedding layers to learn the sign language video frame representation while training \ac{slrt} in an end-to-end manner. However, due to hardware limitations (graphics card memory) we utilize pretrained \acp{cnn} as our spatial embeddings. We extract frame level representations from sign videos and train our sign language transformers to learn \ac{cslr} and \ac{slt} jointly in an end-to-end manner.

In our first set of experiments, we investigate which \ac{cnn} we should be using to represent our sign videos. We utilize state-of-the-art EfficientNets \cite{tan2019efficientnet}, namely B0, B4 and B7, which were trained on ImageNet \cite{deng2009imagenet}. We also use an Inception \cite{szegedy2017inception} network which was pretrained for learning sign language recognition in a CNN+LSTM+HMM setup \cite{koller2019weakly}. In this set of experiments we employed a two layered transformer encoder model.

\begin{table}[!b]
\centering
\resizebox{\columnwidth}{!}{
\begin{tabular}{>{\rowmac}c|>{\rowmac}c>{\rowmac}c|>{\rowmac}c>{\rowmac}c} 
        & \multicolumn{2}{c|}{DEV} & \multicolumn{2}{c}{TEST} \\
Spatial Embedding & del / ins   & WER   & del / ins   & WER   \\ \hline
EfficientNet-B0 & 47.22	/ 1.59 & 57.06 & 46.09 / 1.75 &	56.29 \\
EfficientNet-B4 & 40.73	/ 2.45 & 51.26 & 38.34 / 2.80 &	50.09 \\
EfficientNet-B7 & 39.29 / 2.84 & 50.18 & 37.05 / 2.76 & 47.96 \\
\hline
Pretrained \ac{cnn} & 21.51 / 6.10 & 33.90 & 20.29 / 5.35 &	33.39 \\
\setrow{\bfseries} + BN \& ReLU & 13.54 / 5.74  & 26.70  & 13.85 / 6.43  & 27.62  \clearrow \\
\end{tabular}}
\caption{Impact of the Spatial Embedding Layer variants.}
\label{tbl:slrt:spatial}
\end{table}

Table \ref{tbl:slrt:spatial} shows that as the spatial embedding layer becomes more advanced, \ie B0 vs B7, the recognition performance increases. However, our networks benefited more when we used pretrained features, as these networks had seen sign videos before and learned kernels which can embed more meaningful representations in the latent space. We then tried utilizing Batch Normalization \cite{ioffe2015batch} followed by a ReLU \cite{nair2010rectified} to normalize our inputs and allow our networks to learn more abstract non-linear representations. This improved our results drastically, giving us a boost of nearly 7\% and 6\% of absolute WER reduction on the development and test sets,  respectively. Considering these findings, the rest of our experiments used the batch normalized pretrained \ac{cnn} features of \cite{koller2019weakly} followed by ReLU.

Next, we investigated the effects of having different numbers of transformer layers. Although having a larger number of layers would allow our networks to learn more abstract representations, it also makes them prone to over-fitting. To this end, we built our \ac{slrt} networks using one to six layers and evaluate their \ac{cslr} performance.

\begin{table}[!b]
\centering
\resizebox{0.9\columnwidth}{!}{
\begin{tabular}{>{\rowmac}c|>{\rowmac}c>{\rowmac}c|>{\rowmac}c>{\rowmac}c} 
        & \multicolumn{2}{c|}{DEV} & \multicolumn{2}{c}{TEST} \\
\# Layers  & del/ins   & WER   & del/ins   & WER   \\ \hline
1 & 11.72 / 9.02  & 28.08  & 11.20 / 10.57 & 29.90 \\
2 & 13.54 / 5.74  & 26.70  & 13.85 / 6.43  & 27.62  \\
\setrow{\bfseries} 3 &	11.68 / 6.48  & 24.88  & 11.16 / 6.09  & 24.59 \clearrow \\ 
4 & 12.55 / 5.87  & 24.97  & 13.48 / 6.02 & 26.87 \\
5 & 11.94 /	6.12  & 25.23  & 11.81 / 6.12 & 25.51 \\
6 & 15.01 / 6.11  & 27.46  & 14.30 / 6.28 & 27.78 \\
\end{tabular}}
\caption{Impact of different numbers of layers}
\label{tbl:slrt:layers}
\end{table}

Our recognition performance initially improves with additional layers (See Table~\ref{tbl:slrt:layers}). However, as we continue adding more layers, our networks started to over-fit on the training data, causing performance degradation. In the light of this, for the rest of our experiments, we constructed our sign language transformers using three layers.

\subsection{\textbf{\emph{Sign2Text}} and \textbf{\emph{Sign2(Gloss+Text)}}}
In our next set of experiments we examine the performance gain achieved by unifying the recognition and translation tasks into a single model. As a baseline, we trained a \emph{Sign2Text} network by setting our recognition loss weight $\lambda_\mathrm{R}$ to zero. We then jointly train our sign language transformers, for recognition and translation, with various weightings between the losses.

\begin{table}[!t]
\centering
\resizebox{0.8\columnwidth}{!}{
\begin{tabular}{cc|c|c|c|c} 
\multicolumn{2}{c|}{Loss Weights} & \multicolumn{2}{c|}{DEV} &\multicolumn{2}{c}{TEST} \\
 $\lambda_\mathrm{R}$ & $\lambda_\mathrm{T}$ & WER & BLEU-4 & WER & BLEU-4  \\ \hline
1.0 & 0.0 & 24.88 & -     & 24.59 & -     \\ 
0.0 & 1.0 & -     & 20.69 & -     & 20.17 \\ \hline 
1.0 & 1.0  & 35.13 & 21.73 & 33.75 & 21.22 \\ 
\hline 
2.5 & 1.0  & 26.99 & 22.11 & 27.55 & 21.37 \\ 
5.0 & 1.0  & \textbf{24.61} & 22.12 & \textbf{24.49} & \textbf{21.80} \\ 
10.0 & 1.0 & 24.98 & \textbf{22.38} & 26.16 & 21.32 \\ 
20.0 & 1.0 & 25.87 & 20.90 & 25.73 & 20.93 \\ 
\noalign{\smallskip} 
\end{tabular}
}
\caption{Training Sign Language Transformers to jointly learn recognition and translation with different weight on recognition loss.}
\label{tbl:ph14t:joint}
\vspace{-0.2in}
\end{table}

As can be seen in Table~\ref{tbl:ph14t:joint}, jointly learning recognition and translation with equal weighting ($\lambda_R$=$\lambda_T$=$1.0$) improves the translation performance, while degrading the recognition performance compared to task specific networks. We believe this is due to scale differences of the CTC and word-level cross entropy losses. Increasing the recognition loss weight improved both the recognition and the translation performance, demonstrating the value of sharing training between these related tasks. 

Compared to previously published methods, our Sign Language Transformers surpass both their recognition and translation performance (See Table~\ref{tbl:ph14t:new:baselines}). We report a decrease of 2\% WER over \cite{koller2019weakly} on the test set in both \emph{Sign2Gloss} and \emph{Sign2(Gloss+Text)} setups. More impressively, both our \emph{Sign2Text} and \emph{Sign2(Gloss+Text)} networks doubled the previous state-of-the-art translation results (9.58 vs. 20.17 and 21.32 BLEU-4, respectively). Furthermore, our best performing translation \emph{Sign2(Gloss+Text)} outperforms Camgoz \etal's text-to-text based Gloss2Text translation performance (19.26 vs 21.32 BLEU-4), which was previously proposed as a pseudo upper bound on performance in \cite{camgoz2018neural}. This supports our claim that given more sophisticated network architectures, one would and should achieve better performance translating directly from video representations rather than doing text-to-text translation through a limited gloss representation. 
\vspace{-0.1in}
\section{Qualitative Results}
\label{sec:qual}
In this section we report our qualitative results. We share the spoken language translations generated by our best performing \emph{Sign2(Gloss+Text)} model given sign video representations (See Table~\ref{tbl:qual:results})\footnote{Visit \href{https://github.com/neccam/slt/tree/master/qual}{\textit{our code repository}} for further qualitative examples.}. As the annotations in the \ac{ph14t} dataset are in German, we share both the produced sentences and their translations in English.

\begin{table}[!h]
\vspace{-0.08in}
\centering
\resizebox{\columnwidth}{!}{
    \begin{tabular}{|rl|}
    \hline
    Reference:  & im süden schwacher wind . \\
    ~           & ( in the south gentle wind . ) \\
    Ours:       & der wind weht im süden schwach . \\
    ~           & ( the wind blows gentle in the south . ) \\
    \hline
    \hline
    Reference:  & ähnliches wetter dann auch am donnerstag .  \\
    ~           & ( similar weather then also on thursday . ) \\
    Ours:       &  ähnliches wetter auch am donnerstag . \\
    ~           & ( similar weather also on thursday . ) \\
    \hline
    \hline
    Reference:  & ganz ähnliche temperaturen wie heute zwischen sechs und elf grad . \\
    ~           & ( quite similar temperatures as today between six and eleven degrees . ) \\
    Ours:       & ähnlich wie heute nacht das sechs bis elf grad . \\
    ~           & ( similar as today at night that six to eleven degrees . )\\
    \hline
    \hline
    Reference:  & heute nacht neunzehn bis fünfzehn grad im südosten bis zwölf grad . \\
    ~           & ( tonight nineteen till fifteen degrees in the southeast till twelve degrees . ) \\
    Ours:       & heute nacht werte zwischen neun und fünfzehn grad im südosten bis zwölf grad . \\
    ~           & ( tonight values between nine and fifteen degrees in the southeast till twelve degrees . ) \\
    \hline
    \hline
    Reference:  & am sonntag im norden und in der mitte schauer dabei ist es im norden stürmisch .  \\
    ~           & ( on sunday in the north and center shower while it is stormy in the north . ) \\
    Ours:       & am sonntag im norden und in der mitte niederschläge im norden ist es weiter stürmisch . \\
    ~           & ( on sunday in the north and center rainfall in the north it is continuously stormy . ) \\
    \hline
    \hline
    Reference:  & im süden und südwesten gebietsweise regen sonst recht freundlich . \\
    ~           & ( in the south and southwest partly rain otherwise quite friendly . ) \\
    Ours:       & im südwesten regnet es zum teil kräftig . \\
    ~           & ( in the southwest partly heavy rain . ) \\
    \hline
    \hline
    Reference:  & in der nacht sinken die temperaturen auf vierzehn bis sieben grad . \\
    ~           & ( at night the temperatures lower till fourteen to seven degrees . ) \\
    Ours:       & heute nacht werte zwischen sieben und sieben grad . \\
    ~           & ( tonight values between seven and seven degrees . ) \\
    \hline
    \hline
    Reference:  & heute nacht ist es meist stark bewölkt örtlich regnet oder nieselt es etwas .\\
    ~           & ( tonight it is mostly cloudy locally rain or drizzle . ) \\
    Ours:       &  heute nacht ist es verbreitet wolkenverhangen gebietsweise regnet es kräftig . \\
    ~           & (tonight it is widespread covered with clouds partly strong rain . ) \\
    \hline
    \hline
    Reference:  & an der saar heute nacht milde sechzehn an der elbe teilweise nur acht grad . \\
    ~           & ( at the saar river tonight mild sixteen at the elbe river partly only eight degrees . ) \\
    Ours:       & im rhein und südwesten macht sich morgen nur knapp über null grad . \\
    ~           & ( in the rhine river and south west becomes just above zero degrees . ) \\
    \hline
    \end{tabular}
}
\vspace{-0.08in}
\caption{Generated spoken language translations by our Sign Language Transformers.}
\label{tbl:qual:results}
\vspace{-0.25in}
\end{table}
Overall, the quality of the translations is good, and even where the exact wording differs, it conveys the same information. The most difficult translations seem to be named entities like locations which occur in limited contexts in the training data. Specific numbers are also challenging as there is no grammatical context to distinguish one from another. Despite this, the sentences produced follow standard grammar with surprisingly few exceptions.
\vspace{-0.05in}
\section{Conclusion and Future Work}
\label{sec:conc}

Sign language recognition and understanding is an essential part of the sign language translation task. Previous translation approaches relied heavily on recognition as the initial step of their system. In this paper we proposed Sign Language Transformers, a novel transformer based architecture to jointly learn sign language recognition and translation in an end-to-end manner. We utilized \ac{ctc} loss to inject gloss level supervision into the transformer encoder, training it to do sign language recognition while learning meaningful representations for the end goal of sign language translation, without having an explicit gloss representation as an information bottleneck.

We evaluated our approach on the challenging \ac{ph14t} dataset and report state-of-the-art sign language recognition and translation results, in some cases doubling the performance of previous translation approaches. Our first set of experiments have shown that using features which were pretrained on sign data outperformed using generic ImageNet based spatial representations. Furthermore, we have shown that jointly learning recognition and translation improved the performance across both tasks. More importantly, we have surpassed the text-to-text translation results, which was set as a virtual upper-bound, by directly translating spoken language sentences from video representations. 

As future work, we would like to expand our approach to model multiple sign articulators, namely faces, hands and body, individually to encourage our networks to learn the linguistic relationship between them.


{\small
\section{Acknowledgements} \vspace{-0.1in}
This work received funding from the SNSF Sinergia project `SMILE' (CRSII2\_160811), the European Union's Horizon2020 research and innovation programme under grant agreement no. 762021 `Content4All' and the EPSRC project `ExTOL' (EP/R03298X/1). This work reflects only the author’s view and the Commission is not responsible for any use that may be made of the information it contains. We would also like to thank NVIDIA Corporation for their GPU grant.
}

\clearpage
{\small
\bibliographystyle{ieee_fullname}
\bibliography{Bib/action,Bib/camgoz,Bib/ctc,Bib/deeplearning,Bib/generative,Bib/gesture,Bib/misc,Bib/nmt,Bib/pose,Bib/seq2seq,Bib/sign,Bib/speech}

\begin{thebibliography}{10}\itemsep=-1pt

\bibitem{abadi2016tensorflow}
Mart{\'\i}n Abadi, Paul Barham, Jianmin Chen, Zhifeng Chen, Andy Davis, Jeffrey
  Dean, Matthieu Devin, Sanjay Ghemawat, Geoffrey Irving, Michael Isard,
  Manjunath Kudlur, Josh Levenberg, Rajat Monga, Sherry Moore, Derek~G. Murray,
  Benoit Steiner, Paul Tucker, Vijay Vasudevan, Pete Warden, Martin Wicke, Yuan
  Yu, and Xiaoqiang Zheng.
\newblock {Tensorflow: A System for Large-scale Machine Learning}.
\newblock In {\em Proceedings of the 12th Symposium on Operating Systems Design
  and Implementation}, 2016.

\bibitem{afouras2018deep}
Triantafyllos Afouras, Joon~Son Chung, Andrew Senior, Oriol Vinyals, and Andrew
  Zisserman.
\newblock {Deep Audio-visual Speech Recognition}.
\newblock {\em IEEE Transactions on Pattern Analysis and Machine Intelligence
  (TPAMI)}, 2018.

\bibitem{bahdanau2015neural}
Dzmitry Bahdanau, Kyunghyun Cho, and Yoshua Bengio.
\newblock {Neural Machine Translation by Jointly Learning to Align and
  Translate}.
\newblock In {\em Proceedings of the International Conference on Learning
  Representations (ICLR)}, 2015.

\bibitem{borg2019sign}
Mark Borg and Kenneth~P Camilleri.
\newblock {Sign Language Detection “in the Wild” with Recurrent Neural
  Networks}.
\newblock In {\em Proceedings of the IEEE International Conference on
  Acoustics, Speech, and Signal Processing (ICASSP)}, 2019.

\bibitem{boyes2001hands}
Penny Boyes-Braem and Rachel Sutton-Spence.
\newblock {\em {The Hands are the Head of the Mouth: The Mouth as Articulator
  in Sign Languages}}.
\newblock Gallaudet University Press, 2001.

\bibitem{bragg2019sign}
Danielle Bragg, Oscar Koller, Mary Bellard, Larwan Berke, Patrick Boudrealt,
  Annelies Braffort, Naomi Caselli, Matt Huenerfauth, Hernisa Kacorri, Tessa
  Verhoef, Christian Vogler, and Meredith~Ringel Morris.
\newblock {Sign Language Recognition, Generation, and Translation: An
  Interdisciplinary Perspective}.
\newblock In {\em Proceedings of the International ACM SIGACCESS Conference on
  Computers and Accessibility (ASSETS)}, 2019.

\bibitem{bungeroth2004statistical}
Jan Bungeroth and Hermann Ney.
\newblock {Statistical Sign Language Translation}.
\newblock In {\em Proceedings of the Workshop on Representation and Processing
  of Sign Languages at International Conference on Language Resources and
  Evaluation (LREC)}, 2004.

\bibitem{camgoz2017subunets}
Necati~Cihan Camgoz, Simon Hadfield, Oscar Koller, and Richard Bowden.
\newblock {SubUNets: End-to-end Hand Shape and Continuous Sign Language
  Recognition}.
\newblock In {\em Proceedings of the IEEE International Conference on Computer
  Vision (ICCV)}, 2017.

\bibitem{camgoz2018neural}
Necati~Cihan Camgoz, Simon Hadfield, Oscar Koller, Hermann Ney, and Richard
  Bowden.
\newblock {Neural Sign Language Translation}.
\newblock In {\em Proceedings of the IEEE Conference on Computer Vision and
  Pattern Recognition (CVPR)}, 2018.

\bibitem{camgoz2016sign}
Necati~Cihan Camgoz, Ahmet~Alp Kindiroglu, and Lale Akarun.
\newblock {Sign Language Recognition for Assisting the Deaf in Hospitals}.
\newblock In {\em Proceedings of the International Workshop on Human Behavior
  Understanding (HBU)}, 2016.

\bibitem{camgoz2016bosphorussign}
Necati~Cihan Camgoz, Ahmet~Alp Kindiroglu, Serpil Karabuklu, Meltem Kelepir,
  Ayse~Sumru Ozsoy, and Lale Akarun.
\newblock {BosphorusSign: A Turkish Sign Language Recognition Corpus in Health
  and Finance Domains}.
\newblock In {\em Proceedings of the International Conference on Language
  Resources and Evaluation (LREC)}, 2016.

\bibitem{chai2013sign}
Xiujuan Chai, Guang Li, Yushun Lin, Zhihao Xu, Yili Tang, Xilin Chen, and Ming
  Zhou.
\newblock {Sign Language Recognition and Translation with Kinect}.
\newblock In {\em Proceedings of the International Conference on Automatic Face
  and Gesture Recognition (FG)}, 2013.

\bibitem{cherniavsky2008activity}
Neva Cherniavsky, Richard~E Ladner, and Eve~A Riskin.
\newblock {Activity Detection in Conversational Sign Language Video for Mobile
  Telecommunication}.
\newblock In {\em Proceedings of the International Conference on Automatic Face
  and Gesture Recognition (FG)}, 2008.

\bibitem{cooper2009learning}
Helen Cooper and Richard Bowden.
\newblock {Learning Signs from Subtitles: A Weakly Supervised Approach to Sign
  Language Recognition}.
\newblock In {\em Proceedings of the IEEE Conference on Computer Vision and
  Pattern Recognition (CVPR)}, 2009.

\bibitem{cooper2011sign}
Helen Cooper, Brian Holt, and Richard Bowden.
\newblock {Sign Language Recognition}.
\newblock In {\em Visual Analysis of Humans}. Springer, 2011.

\bibitem{cormier2019extol}
Kearsy Cormier, Neil Fox, Bencie Woll, Andrew Zisserman, {Necati Cihan} Camgoz,
  and Richard Bowden.
\newblock {ExTOL: Automatic recognition of British Sign Language using the BSL
  Corpus}.
\newblock In {\em Proceedings of the Sign Language Translation and Avatar
  Technology (SLTAT)}, 2019.

\bibitem{cui2017recurrent}
Runpeng Cui, Hu Liu, and Changshui Zhang.
\newblock {Recurrent Convolutional Neural Networks for Continuous Sign Language
  Recognition by Staged Optimization}.
\newblock In {\em Proceedings of the IEEE Conference on Computer Vision and
  Pattern Recognition (CVPR)}, 2017.

\bibitem{cui2019deep}
Runpeng Cui, Hu Liu, and Changshui Zhang.
\newblock {A Deep Neural Framework for Continuous Sign Language Recognition by
  Iterative Training}.
\newblock {\em IEEE Transactions on Multimedia}, 2019.

\bibitem{dai2019transformer}
Zihang Dai, Zhilin Yang, Yiming Yang, William~W Cohen, Jaime Carbonell, Quoc~V
  Le, and Ruslan Salakhutdinov.
\newblock {Transformer-XL: Attentive Language Models beyond a Fixed-length
  Context}.
\newblock In {\em Proceedings of the Annual Meeting of the Association for
  Computational Linguistics (ACL)}, 2019.

\bibitem{deng2009imagenet}
Jia Deng, Wei Dong, Richard Socher, Li-Jia Li, Kai Li, and Fei-Fei Li.
\newblock {ImageNet: A Large-scale Hierarchical Image Database}.
\newblock In {\em Proceedings of the IEEE Conference on Computer Vision and
  Pattern Recognition (CVPR)}, 2009.

\bibitem{devlin2018bert}
Jacob Devlin, Ming-Wei Chang, Kenton Lee, and Kristina Toutanova.
\newblock {BERT: Pre-training of Deep Bidirectional Transformers for Language
  Understanding}.
\newblock In {\em Proceedings of the Conference of the North {A}merican Chapter
  of the Association for Computational Linguistics (ACL): Human Language
  Technologies}, 2019.

\bibitem{ebling2016Automatic}
Sarah Ebling.
\newblock {\em {Automatic Translation from German to Synthesized Swiss German
  Sign Language}}.
\newblock PhD thesis, University of Zurich, 2016.

\bibitem{ebling2018smile}
Sarah Ebling, Necati~Cihan Camgoz, Penny~Boyes Braem, Katja Tissi, Sandra
  Sidler-Miserez, Stephanie Stoll, Simon Hadfield, Tobias Haug, Richard Bowden,
  Sandrine Tornay, Marzieh Razavi, and Mathew Magimai-Doss.
\newblock {SMILE Swiss German Sign Language Dataset}.
\newblock In {\em Proceedings of the International Conference on Language
  Resources and Evaluation (LREC)}, 2018.

\bibitem{ebling2015bridging}
Sarah Ebling and Matt Huenerfauth.
\newblock {Bridging the Gap between Sign Language Machine Translation and Sign
  Language Animation using Sequence Classification}.
\newblock In {\em Proceedings of the 6th Workshop on Speech and Language
  Processing for Assistive Technologies (SPLAT)}, 2015.

\bibitem{forster2012rwth}
Jens Forster, Christoph Schmidt, Thomas Hoyoux, Oscar Koller, Uwe Zelle,
  Justus~H Piater, and Hermann Ney.
\newblock {RWTH-PHOENIX-Weather: A Large Vocabulary Sign Language Recognition
  and Translation Corpus}.
\newblock In {\em Proceedings of the International Conference on Language
  Resources and Evaluation (LREC)}, 2012.

\bibitem{forster2014extensions}
Jens Forster, Christoph Schmidt, Oscar Koller, Martin Bellgardt, and Hermann
  Ney.
\newblock {Extensions of the Sign Language Recognition and Translation Corpus
  RWTH-PHOENIX-Weather}.
\newblock In {\em Proceedings of the International Conference on Language
  Resources and Evaluation (LREC)}, 2014.

\bibitem{gehring2017convolutional}
Jonas Gehring, Michael Auli, David Grangier, Denis Yarats, and Yann~N Dauphin.
\newblock {Convolutional Sequence to Sequence Learning}.
\newblock In {\em Proceedings of the ACM International Conference on Machine
  Learning (ICML)}, 2017.

\bibitem{glorot2010understanding}
Xavier Glorot and Yoshua Bengio.
\newblock {Understanding the Difficulty of Training Deep Feedforward Neural
  Networks}.
\newblock In {\em Proceedings of the International Conference on Artificial
  Intelligence and Statistics}, 2010.

\bibitem{goodfellow2016deep}
Ian Goodfellow, Yoshua Bengio, and Aaron Courville.
\newblock {\em {Deep Learning}}.
\newblock MIT press, 2016.

\bibitem{graves2006connectionist}
Alex Graves, Santiago Fern{\'a}ndez, Faustino Gomez, and J{\"u}rgen
  Schmidhuber.
\newblock {Connectionist Temporal Classification: Labelling Unsegmented
  Sequence Data with Recurrent Neural Networks}.
\newblock In {\em Proceedings of the ACM International Conference on Machine
  Learning (ICML)}, 2006.

\bibitem{hanke2010dgs}
Thomas Hanke, Lutz K{\"o}nig, Sven Wagner, and Silke Matthes.
\newblock {DGS Corpus \& Dicta-Sign: The Hamburg Studio Setup}.
\newblock In {\em Proceedings of the Representation and Processing of Sign
  Languages: Corpora and Sign Language Technologies}, 2010.

\bibitem{huang2018video}
Jie Huang, Wengang Zhou, Qilin Zhang, Houqiang Li, and Weiping Li.
\newblock {Video-based Sign Language Recognition without Temporal
  Segmentation}.
\newblock In {\em Proceedings of the AAAI Conference on Artificial
  Intelligence}, 2018.

\bibitem{ioffe2015batch}
Sergey Ioffe and Christian Szegedy.
\newblock {Batch Normalization: Accelerating Deep Network Training by Reducing
  Internal Covariate Shift}.
\newblock In {\em Proceedings of the International Conference on Machine
  Learning (ICML)}, 2015.

\bibitem{irie2019language}
Kazuki Irie, Albert Zeyer, Ralf Schl{\"u}ter, and Hermann Ney.
\newblock {Language Modeling with Deep Transformers}.
\newblock In {\em 20th Annual Conference of the International Speech
  Communication Association (INTERSPEECH)}, 2019.

\bibitem{joze2019ms}
Hamid Reza~Vaezi Joze and Oscar Koller.
\newblock {MS-ASL: A Large-Scale Data Set and Benchmark for Understanding
  American Sign Language}.
\newblock In {\em Proceedings of the British Machine Vision Conference (BMVC)},
  2019.

\bibitem{khan2014pause}
Shujjat Khan, Donald~G Bailey, and Gourab~Sen Gupta.
\newblock {Pause detection in continuous sign language}.
\newblock {\em International Journal of Computer Applications in Technology},
  50, 2014.

\bibitem{kingma2014adam}
Diederik~P. Kingma and Jimmy Ba.
\newblock {Adam: A Method for Stochastic Optimization}.
\newblock In {\em Proceedings of the International Conference on Learning
  Representations (ICLR)}, 2014.

\bibitem{ko2019neural}
Sang-Ki Ko, Chang~Jo Kim, Hyedong Jung, and Choongsang Cho.
\newblock {Neural Sign Language Translation based on Human Keypoint
  Estimation}.
\newblock {\em Applied Sciences}, 9(13), 2019.

\bibitem{koller2019weakly}
Oscar Koller, Necati~Cihan Camgoz, Richard Bowden, and Hermann Ney.
\newblock {Weakly Supervised Learning with Multi-Stream CNN-LSTM-HMMs to
  Discover Sequential Parallelism in Sign Language Videos}.
\newblock {\em {IEEE Transactions on Pattern Analysis and Machine Intelligence
  (TPAMI)}}, 2019.

\bibitem{koller2015continuous}
Oscar Koller, Jens Forster, and Hermann Ney.
\newblock {Continuous Sign Language Recognition: Towards Large Vocabulary
  Statistical Recognition Systems Handling Multiple Signers}.
\newblock {\em Computer Vision and Image Understanding (CVIU)}, 141, 2015.

\bibitem{koller2017resign}
Oscar Koller, Sepehr Zargaran, and Hermann Ney.
\newblock {Re-Sign: Re-Aligned End-to-End Sequence Modelling with Deep
  Recurrent CNN-HMMs}.
\newblock In {\em Proceedings of the IEEE Conference on Computer Vision and
  Pattern Recognition (CVPR)}, 2017.

\bibitem{koller2016deepsign}
Oscar Koller, Sepehr Zargaran, Hermann Ney, and Richard Bowden.
\newblock {Deep Sign: Hybrid CNN-HMM for Continuous Sign Language Recognition}.
\newblock In {\em Proceedings of the British Machine Vision Conference (BMVC)},
  2016.

\bibitem{kreutzer2019joey}
Julia Kreutzer, Joost Bastings, and Stefan Riezler.
\newblock Joey {NMT}: A minimalist {NMT} toolkit for novices.
\newblock In {\em Proceedings of the Conference on Empirical Methods in Natural
  Language Processing (EMNLP): System Demonstrations}, 2019.

\bibitem{luong2015effective}
Minh-Thang Luong, Hieu Pham, and Christopher~D Manning.
\newblock {Effective Approaches to Attention-based Neural Machine Translation}.
\newblock In {\em Proceedings of the Conference on Empirical Methods in Natural
  Language Processing (EMNLP)}, 2015.

\bibitem{morrissey2008data}
Sara Morrissey.
\newblock {\em {Data-driven Machine Translation for Sign Languages}}.
\newblock PhD thesis, Dublin City University, 2008.

\bibitem{morrissey2010building}
Sara Morrissey, Harold Somers, Robert Smith, Shane Gilchrist, and Sandipan
  Dandapat.
\newblock {Building a Sign Language Corpus for Use in Machine Translation}.
\newblock In {\em Proceedings of the Representation and Processing of Sign
  Languages: Corpora and Sign Language Technologies}, 2010.

\bibitem{nair2010rectified}
Vinod Nair and Geoffrey~E Hinton.
\newblock {Rectified Linear Units Improve Restricted Boltzmann Machines}.
\newblock In {\em Proceedings of the International Conference on Machine
  Learning (ICML)}, 2010.

\bibitem{neubig2017neural}
Graham Neubig.
\newblock {Neural Machine Translation and Sequence-to-Sequence Models: A
  Tutorial}.
\newblock {\em arXiv:1703.01619}, 2017.

\bibitem{papineni2002bleu}
Kishore Papineni, Salim Roukos, Todd Ward, and Wei-Jing Zhu.
\newblock {BLEU: A Method for Automatic Evaluation of Machine Translation}.
\newblock In {\em Proceedings of the Annual Meeting of the Association for
  Computational Linguistics (ACL)}, 2002.

\bibitem{paszke2017automatic}
Adam Paszke, Sam Gross, Soumith Chintala, Gregory Chanan, Edward Yang, Zachary
  DeVito, Zeming Lin, Alban Desmaison, Luca Antiga, and Adam Lerer.
\newblock {Automatic Differentiation in PyTorch}.
\newblock In {\em Proceedings of the Advances in Neural Information Processing
  Systems Workshops (NIPSW)}, 2017.

\bibitem{sandler1993sign}
Wendy Sandler.
\newblock {Sign Language and Modularity}.
\newblock {\em Lingua}, 89(4), 1993.

\bibitem{santemiz2009automatic}
Pinar Santemiz, Oya Aran, Murat Saraclar, and Lale Akarun.
\newblock {Automatic Sign Segmentation from Continuous Signing via Multiple
  Sequence Alignment}.
\newblock In {\em Proceedings of the IEEE International Conference on Computer
  Vision Workshops (ICCVW)}, 2009.

\bibitem{schembri2013building}
Adam Schembri, Jordan Fenlon, Ramas Rentelis, Sally Reynolds, and Kearsy
  Cormier.
\newblock {Building the British Sign Language Corpus}.
\newblock {\em Language Documentation \& Conservation (LD\&C)}, 7, 2013.

\bibitem{schmid2013using}
Christoph Schmidt, Oscar Koller, Hermann Ney, Thomas Hoyoux, and Justus Piater.
\newblock {Using Viseme Recognition to Improve a Sign Language Translation
  System}.
\newblock In {\em Proceedings of the International Workshop on Spoken Language
  Translation}, 2013.

\bibitem{shipman2017speed}
Frank~M Shipman, Satyakiran Duggina, Caio~DD Monteiro, and Ricardo
  Gutierrez-Osuna.
\newblock {Speed-Accuracy Tradeoffs for Detecting Sign Language Content in
  Video Sharing Sites}.
\newblock In {\em Proceedings of the International ACM SIGACCESS Conference on
  Computers and Accessibility (ASSETS)}, 2017.

\bibitem{starner1998real}
Thad Starner, Joshua Weaver, and Alex Pentland.
\newblock {Real-time American Sign Language Recognition using Desk and Wearable
  Computer Based Video}.
\newblock {\em IEEE Transactions on Pattern Analysis and Machine Intelligence
  (TPAMI)}, 20(12), 1998.

\bibitem{stein2012analysis}
Daniel Stein, Christoph Schmidt, and Hermann Ney.
\newblock {Analysis, Preparation, and Optimization of Statistical Sign Language
  Machine Translation}.
\newblock {\em Machine Translation}, 26(4), 2012.

\bibitem{stokoe1980sign}
William~C Stokoe.
\newblock {Sign Language Structure}.
\newblock {\em Annual Review of Anthropology}, 9(1), 1980.

\bibitem{stoll2018sign}
Stephanie Stoll, Necati~Cihan Camgoz, Simon Hadfield, and Richard Bowden.
\newblock {Sign Language Production using Neural Machine Translation and
  Generative Adversarial Networks}.
\newblock In {\em Proceedings of the British Machine Vision Conference (BMVC)},
  2018.

\bibitem{stoll2020text2sign}
Stephanie Stoll, Necati~Cihan Camgoz, Simon Hadfield, and Richard Bowden.
\newblock {Text2Sign: Towards Sign Language Production using Neural Machine
  Translation and Generative Adversarial Networks}.
\newblock {\em {International Journal of Computer Vision (IJCV)}}, 2020.

\bibitem{sutskever2014sequence}
Ilya Sutskever, Oriol Vinyals, and Quoc~V Le.
\newblock {Sequence to Sequence Learning with Neural Networks}.
\newblock In {\em Proceedings of the Advances in Neural Information Processing
  Systems (NIPS)}, 2014.

\bibitem{sutton1999linguistics}
Rachel Sutton-Spence and Bencie Woll.
\newblock {\em {The Linguistics of British Sign Language: An Introduction}}.
\newblock Cambridge University Press, 1999.

\bibitem{suzgun2015hospisign}
Muhammed~Mirac Suzgun, Hilal Ozdemir, Necati~Cihan Camgoz, Ahmet Kindiroglu,
  Dogac Basaran, Cengiz Togay, and Lale Akarun.
\newblock {Hospisign: An Interactive Sign Language Platform for Hearing
  Impaired}.
\newblock In {\em Proceedings of the International Conference on Computer
  Graphics, Animation and Gaming Technologies (Eurasia Graphics)}, 2015.

\bibitem{szegedy2017inception}
Christian Szegedy, Sergey Ioffe, Vincent Vanhoucke, and Alexander~A Alemi.
\newblock {Inception-v4, Inception-ResNet and the Impact of Residual
  Connections on Learning}.
\newblock In {\em Proceedings of the AAAI Conference on Artificial
  Intelligence.}, 2017.

\bibitem{tamura1988recognition}
Shinichi Tamura and Shingo Kawasaki.
\newblock {Recognition of Sign Language Motion Images}.
\newblock {\em Pattern Recognition}, 21(4), 1988.

\bibitem{tan2019efficientnet}
Mingxing Tan and Quoc~V Le.
\newblock {EfficientNet: Rethinking Model Scaling for Convolutional Neural
  Networks}.
\newblock In {\em Proceedings of the International Conference on Machine
  Learning (ICML)}, 2019.

\bibitem{tornay2019hmm}
Sandrine Tornay, Marzieh Razavi, Necati~Cihan Camgoz, Richard Bowden, and
  Mathew Magimai-Doss.
\newblock {HMM-based Approaches to Model Multichannel Information in Sign
  Language Inspired from Articulatory Features-based Speech Processing}.
\newblock In {\em Proceedings of the IEEE International Conference on
  Acoustics, Speech, and Signal Processing (ICASSP)}, 2019.

\bibitem{tsai2019MULT}
Yao-Hung~Hubert Tsai, Shaojie Bai, Paul~Pu Liang, , J.~Zico Kolter,
  Louis-Philippe Morency, and Ruslan Salakhutdinov.
\newblock {Multimodal Transformer for Unaligned Multimodal Language Sequences}.
\newblock In {\em Proceedings of the Annual Meeting of the Association for
  Computational Linguistics (ACL)}, 2019.

\bibitem{van1991automatic}
Jan~P van Hemert.
\newblock {Automatic Segmentation of Speech}.
\newblock {\em IEEE Transactions on Signal Processing}, 39(4), 1991.

\bibitem{vaswani2017attention}
Ashish Vaswani, Noam Shazeer, Niki Parmar, Jakob Uszkoreit, Llion Jones,
  Aidan~N Gomez, {\L}ukasz Kaiser, and Illia Polosukhin.
\newblock {Attention is All You Need}.
\newblock In {\em Proceedings of the Advances in Neural Information Processing
  Systems (NIPS)}, 2017.

\bibitem{wang2016sparse}
Hanjie Wang, Xiujuan Chai, and Xilin Chen.
\newblock {Sparse Observation (SO) Alignment for Sign Language Recognition}.
\newblock {\em Neurocomputing}, 175, 2016.

\bibitem{wang2016grassman}
Hanjie Wang, Xiujuan Chai, Xiaopeng Hong, Guoying Zhao, and Xilin Chen.
\newblock {Isolated Sign Language Recognition with Grassmann Covariance
  Matrices}.
\newblock {\em ACM Transactions on Accessible Computing}, 8(4), 2016.

\bibitem{wang2018non}
Xiaolong Wang, Ross Girshick, Abhinav Gupta, and Kaiming He.
\newblock Non-local neural networks.
\newblock In {\em Proceedings of the IEEE Conference on Computer Vision and
  Pattern Recognition (CVPR)}, pages 7794--7803, 2018.

\bibitem{wu2016google}
Yonghui Wu, Mike Schuster, Zhifeng Chen, Quoc~V. Le, Mohammad Norouzi, Wolfgang
  Macherey, Maxim Krikun, Yuan Cao, Qin Gao, Klaus Macherey, Jeff Klingner,
  Apurva Shah, Melvin Johnson, Xiaobing Liu, Łukasz Kaiser, Stephan Gouws,
  Yoshikiyo Kato, Taku Kudo, Hideto Kazawa, Keith Stevens, George Kurian,
  Nishant Patil, Wei Wang, Cliff Young, Jason Smith, Jason Riesa, Alex Rudnick,
  Oriol Vinyals, Greg Corrado, Macduff Hughes, and Jeffrey Dean.
\newblock {Google's Neural Machine Translation System: Bridging the Gap Between
  Human and Machine Translation}.
\newblock {\em arXiv:1609.08144}, 2016.

\bibitem{yin2016iterative}
Fang Yin, Xiujuan Chai, and Xilin Chen.
\newblock {Iterative Reference Driven Metric Learning for Signer Independent
  Isolated Sign Language Recognition}.
\newblock In {\em Proceedings of the European Conference on Computer Vision
  (ECCV)}, 2016.

\bibitem{yoshida2002automatic}
Norimasa Yoshida.
\newblock {\em {Automatic Utterance Segmentation in Spontaneous Speech}}.
\newblock PhD thesis, Massachusetts Institute of Technology, 2002.

\bibitem{zhang2019ernie}
Zhengyan Zhang, Xu Han, Zhiyuan Liu, Xin Jiang, Maosong Sun, and Qun Liu.
\newblock {ERNIE: Enhanced Language Representation with Informative Entities}.
\newblock In {\em Proceedings of the Annual Meeting of the Association for
  Computational Linguistics (ACL)}, 2019.

\end{thebibliography}
}

\end{document}